%% file: main.tex
\documentclass[sigconf]{acmart}

\usepackage{booktabs} 
\setcopyright{rightsretained}
\usepackage{epigraph}
\usepackage{color}

\acmDOI{}

\acmISBN{}

\acmConference[ML4Creativity'17]{Workshop on Machine Learning for Creativity, SIGKDD}{August 2017}{Nova Scotia, Canada} 
\acmYear{2017}
\copyrightyear{2017}

\acmPrice{}

\graphicspath{{./images/pdf/}}
\begin{document}
\title{"Let me convince you to buy my product ... "}
\subtitle{A Case Study of an Automated Persuasive System for Fashion Products}

\author{Vitobha Munigala, Srikanth Tamilselvam, Anush Sankaran}
\affiliation{%
  \institution{IBM Research, India}
}
\email{{vmunigal, srikanth.tamilselvam, anussank}@in.ibm.com}

\renewcommand{\shortauthors}{Vitobha et al.}

\begin{abstract}
Persuasivenes is a creative art aimed at making people believe in certain set of beliefs. Many a times, such creativity is about adapting richness of one domain into another to strike
a chord with the target audience. In this research, we present PersuAIDE! - A persuasive system based on linguistic creativity to transform given sentence to generate various forms of persuading sentences. These various forms cover multiple focus of persuasion such as memorability and sentiment. For a given simple product line, the algorithm is composed of several steps including: (i) select an appropriate well-known expression for the target domain to add memorability, (ii) identify keywords and entities in the given sentence and expression and transform it to produce creative persuading sentence, and (iii) adding positive or negative sentiment for further persuasion. The persuasive conversion were manually verified using qualitative results and the effectiveness of the proposed approach is empirically discussed.
\end{abstract}

%
%



\maketitle

\input{1_introduction}
\input{2_Literature}
\input{3_Proposed_Approach}
\input{4_Experimental_Study}
\input{5_Working_System}

\input{7_Conclusion}

\bibliographystyle{ACM-Reference-Format}
\bibliography{sigproc} 

\end{document}

%% file: 1_introduction.tex
\section{Introduction}

\begin{figure}[!t]
	\centering
	\includegraphics[width=3.4in]{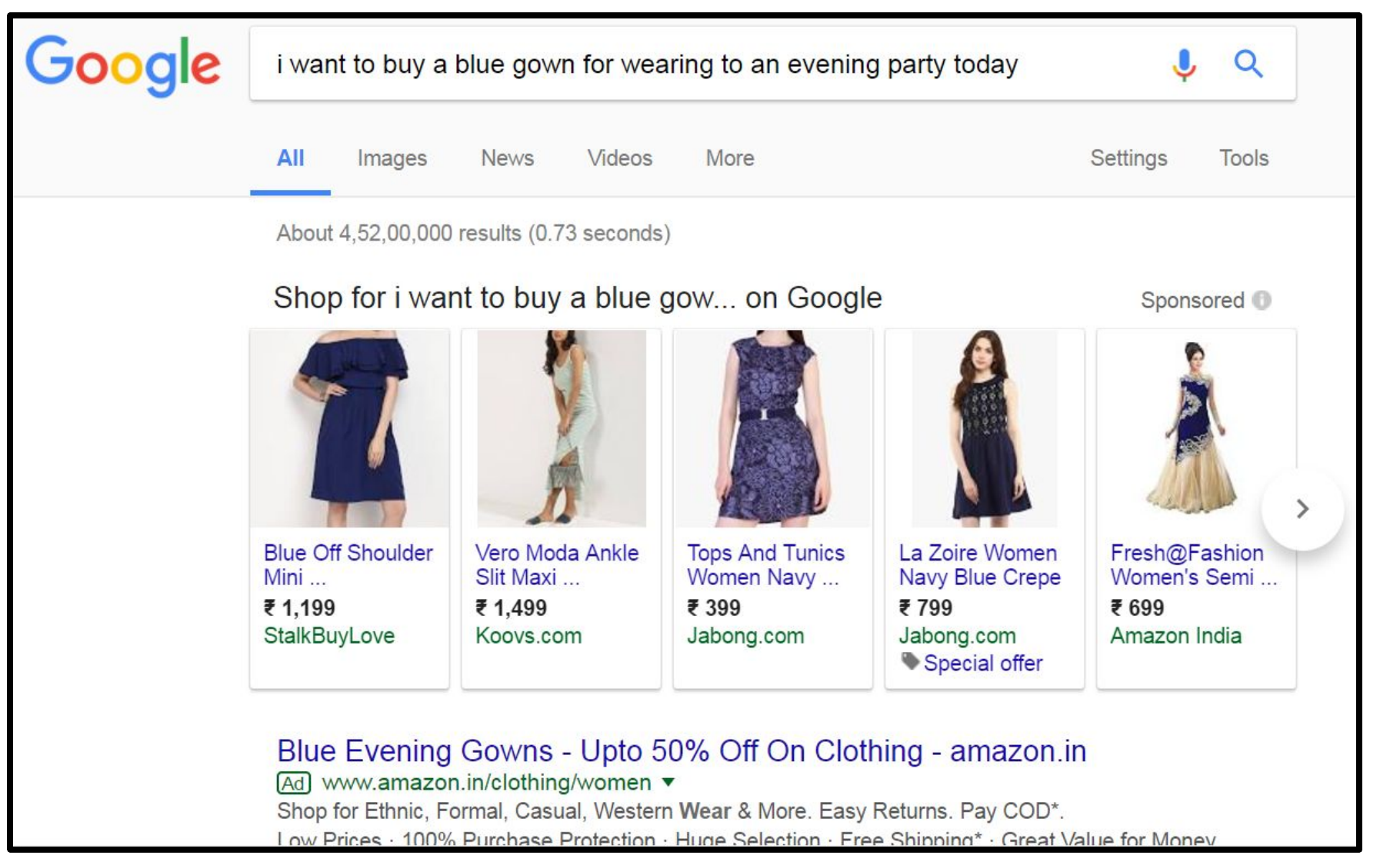}
	\caption{Summarized results obtained from the Google search for an input query, ``I want to buy a blue gown for wearing to an evening party".}
	\label{fig:google}
\end{figure}

\begin{figure}[ht]
	\centering
	\includegraphics[width=3.4in]{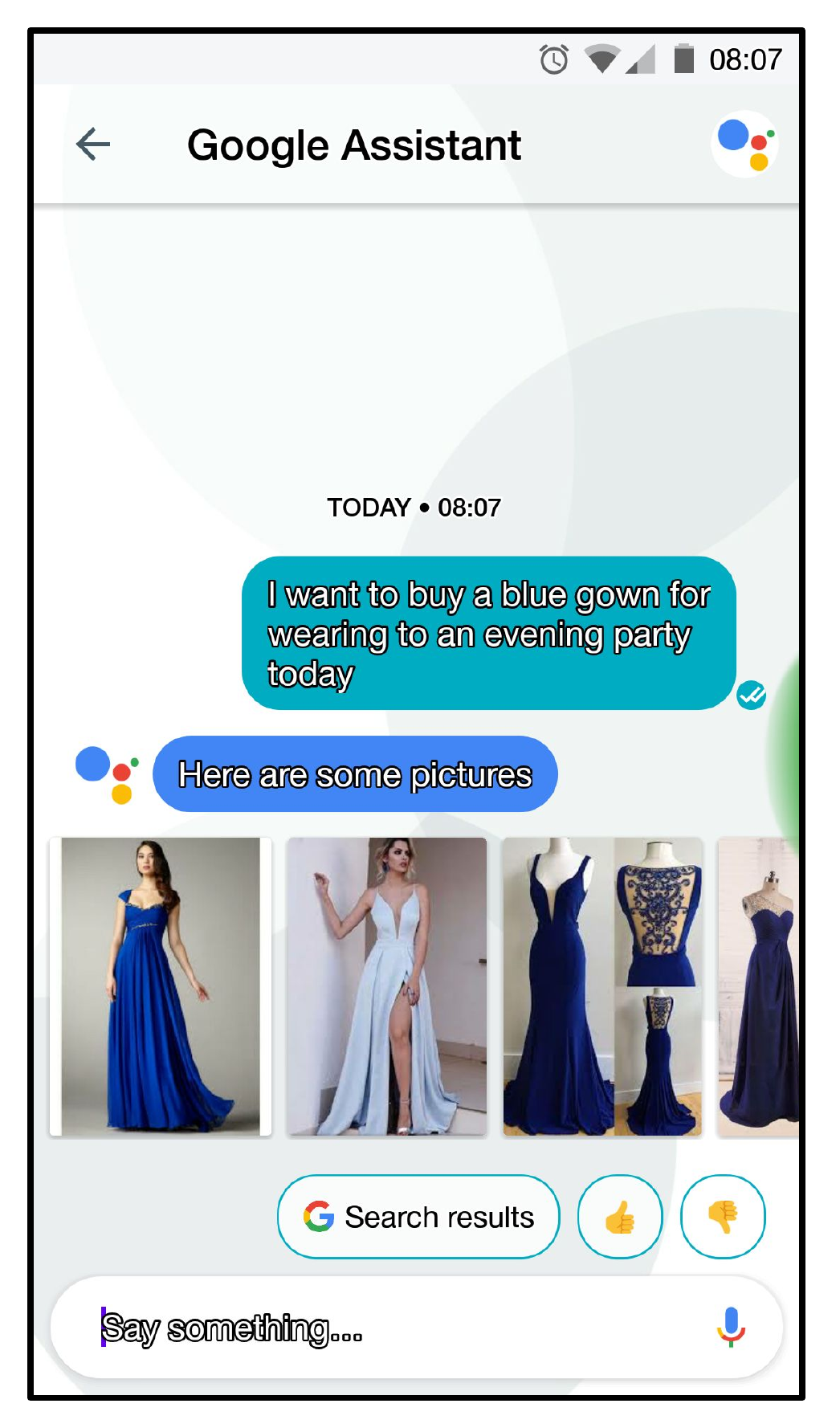}
	\caption{Personalized results obtained from the Google's assistant, Allo for an input query, ``I want to buy a blue gown for wearing to an evening party".}
	\label{fig:allo}
\end{figure}

Consider the following question, ``I want to buy a blue gown for wearing to an evening party?". The cognitive search systems, could find and retrieve all the relevant ``\textit{blue gown}" dresses that are popularly voted as a suitable fashion wear for an ``\textit{evening party}". The state-of-art systems could summarize content from various source and present in a much readable manner. The summarized response from Google's search is shown in Figure~\ref{fig:google}. 
More so, the systems could even point to the most relevant e-commerce websites where the dress could be purchased. Figure~\ref{fig:allo} shows the results from a personalized mobile assistant, Google's Allo, where each image is an hyperlink to it's corresponding e-commerce website. 
Existing question-answering systems and conversation systems are well trained for resolving the facts from the input query and finding semantic similarity with respect to a huge corpus of background dataset.

The retrieved summarized results can be observed to provide descriptions such as ``Blue Off Shoulder Mini Skater Dress", ``Vero Moda Ankle Slit Maxi Dress". While these descriptions are factually relevant to both the dress as well as the input query, it does not persuade or compel me to purchase the corresponding product. In contradiction, a description such as ``The dress choses its owner! This gleaming off shoulder mini skater would create a magical attraction in your evening party", would make this product more catchy, memorable, and compelling to buy.
The motivation for our research is derived from this existing gap - Could there be a creative system that could persuade me and sell me a product, rather than just describing the product?. We propose a system, \textit{PersuAIDE!} that better persuades the consumer to sell a particular product. 

This kind of persuasion is an common technique employed by sales people to enhance the impression of their products in the face of their consumers. Like in a caricature, this process involves highlighting and amplifying a few important facts that could be personalized for the consumer, while also inhibiting (or down-playing) the remaining facts that is not relevant to persuade the consumer. Such a creative feature could be incorporated into any search, question-answering, or conversation system and also could be used for targeted marketing in any e-commerce business. While the applications could be broad and far-fetching, to have a specific focus in this research, we have taken a case study in the fashion domain to persuade and sell fashionable couture and relevant accessories.

The problem statement is formulated as, given a bland product description sentence about a fashion product, transform the sentence to be able to better persuade the consumer. On a broad level, such a problem statement could be compared with a sentence paraphrasing task~\cite{37566}~\cite{serban2016generating}, we have focused on some specific aspects of persuasion, which cannot be directly incorporated into a sentence rephrasing setting,
\begin{enumerate}
	\item Memorability~\cite{guerini2015echoes}: Plain descriptions and bland facts do not tend to be attract the consumers. To increase the memorability of such descriptions, popular slogans, movie titles/ dialogue, poetry, catchy-phrases could be adopted to suit the product's description.
	\item Sentiment~\cite{pang2008opinion}: Addition of positive sentiments in the description could influence the consumer's inclination towards the product. One such way of adding sentiments into sentences could be to add enhancing adjectives to noun-phrases in the sentence. For example, a description such as ``Blue Embrodered Net Semi-Stitched Anarkali Gown" could be enhanced as ``\textit{Dazzling} Blue Embrodered Net Semi-Stitched Anarkali \textit{Gracious} Gown"
\end{enumerate}

In the rest of the paper, Section 2 briefs the closely related literature work, Section 3 explains the proposed approach, Section 4 discusses the obtained results, Section 5 details some of the future extensions of this work, and Section 6 concludes our research.

%% file: 2_Literature.tex
\section{Literature Study}
Usage of linguistic creativity in the past have been looked into for poetry generation. \cite{toivanen2012corpus} proposed a poetry generation system that uses morphological analysis to identify substitutable words for the given topic to generate variants of poem. Recently, there is lot of focus on creative text generation based on lexical substitution. \cite{guerini2011slanting} discussed about slanting an original expression to more positive or negative version by adding or subtracting a word. \cite{lynchevery} discussed about identifying various topic in the news and choosing the right expression that captures them. \cite{ozbal2013brainsup} proposed an extensible framework that allows users to force multiple words to appear in the expression. Given keywords, the framework comes up with an expression that includes them. More recently, \cite{gatti2015slogans} proposed a system that blends expression with best matching evolving news. \cite{gatti2016heady} proposed a creative system that generates a news headlines based on best matching expression. Almost all of these work could be considered as catering to memorability aspect of persuasion alone as they rely on popular expression, quote, poem etc. and their input text are fairly expressive text like headline which are fairly easy to persuade compared to product description which are blunt statements. Also none of them discussed about transforming sentences to include complex persuasive features like argumentation, sentiment etc in the target domain. 

%% file: 3_Proposed_Approach.tex
\section{PersuAIDE!: Proposed Approach}

\begin{figure}[htb]
	\centering
	\includegraphics[width=0.42\textwidth]{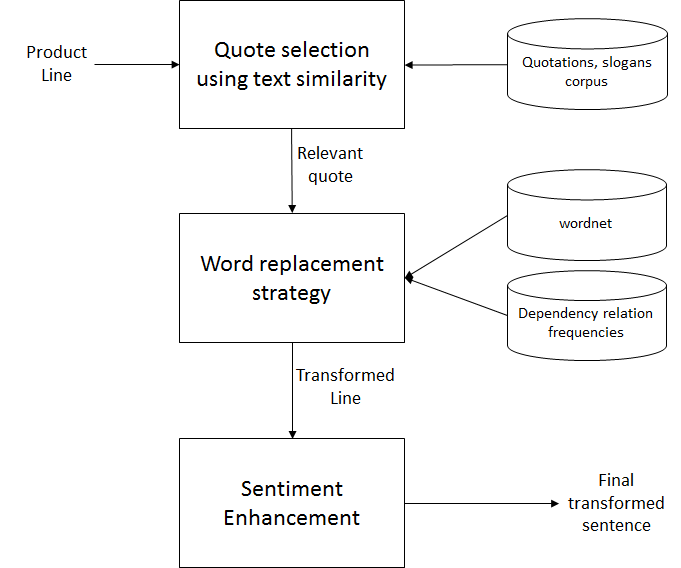}
	\caption{\label{fig:sys_arch}An overview of sentence transformation}
\end{figure}

In this section, we discuss the details of the proposed persuasive system. Figure \ref{fig:sys_arch} shows an overview of the proposed system. Further explanation of multi-step transformation process is as follows,\\

\noindent \textbf{Step 1:} From a large corpus of text data, capture fashion domain knowledge in terms of dependency relationship between different words commonly used in fashion. In order to create this domain knowledge, we collected a large corpus of fashion related text from some of the leading fashion magazines: Chictopia\footnote{http://www.chictopia.com/} and Vogue fashion~\footnote{http://www.vogue.in/fashion/}. From these blog posts and magazine articles, we extracted about $425,000$ text sentences. This dataset is appended with a list of popular slogans, cliches, movie dialogs, quotations~\cite{gatti2015slogans} to find the prominent dependency relations in fashion domain. We use the Stanford's dependency parser~\cite{de2014universal} to get the dependency relation in each sentence of the training corpus and store the frequency of each such relation. We obtained a dependency relationship matrix whose rows and columns are the words in vocabulary and the integer in each cell $f(a, b)$ denotes the frequency of a relation (between row words and the column word) in the entire corpus. In our dataset, we had a vocabulary size of $51,000$ words and $4,600,000$ dependency relations.\\
	
\noindent \textbf{Step 2:} This step is to improve the memorability of the product description. Once the dependency relations are captured, we get to the process of sentence transformation. Given a sentence, we now transform it into another form by making use of quotations, slogans, etc while taking into consideration the domain dependency relations. For a given product description, we find the most similar auxiliary sentence from the quotations, slogans, movie dialogs corpus with the help of an LSTM based text similarity algorithm. \\
	
\noindent \textbf{Step 3:} Once the auxiliary sentence is identified, the sentence transformation can happen in two ways. Either by substituting few words in the product description sentence using the words in the auxiliary sentence or introducing new words into the auxiliary sentence from the words in the product description sentence. \\

\noindent \textbf{Step 4:} For word substitution, we capture all the nouns, verbs, adjectives and adverbs in the input and auxiliary sentences. For each of the selected word from the auxiliary sentence, we find its derivationally related forms with the help of wordnet. These derivationally related forms are bucketed into four categories (nouns, verbs, adjectives, adverbs) based on their POS tags. Now, for each of the selected word from input sentence, we choose the best substituting word from one of the created buckets. The bucket is chosen based on the POS tag of the input word. In the chosen bucket, we find the best word by making use of the following procedure.
\begin{itemize}
	\item Find all the dependency relations from the input sentence which involves the chosen word `w'.
	\item For each word `k'in the chosen bucket, we find the matching score to replace the word w. The matching score is defines as below.
	\begin{equation}
	score(R, w, k) = exp\left(\sum_{R \ni w} \frac{\log f(a, b)}{|R \ni w|}\right)
	\end{equation}
	
	where, \textit{R} denotes the dependency relations in the input sentence and \textit{f(a, b)} denotes the frequency of the dependency relation $a \to b$ in the training corpus, and $R \ni w$ denotes the set of all relations containing the word $w$. If $w$ is child in the dependency relation $r$ then, $a$ = parent in the relation $r$ and $b$ = $k$. If $w$ is parent in the dependency relation $r$ then, $a$ = $k$ and $b$ = parent in the relation $r$.
	For each selected word in the input sentence, we select $k$ which has maximum $matching\ score$ and replace the word $w$. 
\end{itemize}

\noindent \textbf{Step 5:} For word addition, we select word from a list of positive sense adjectives like $wonderful$, $wondeful$, $dazzling$, etc. For each noun in the input sentence, we check the possibility of introducing a suitable adjective. In fashion text, we observed that most of the nouns are colors, garments (outfits), jewelry, etc. Based on the type of noun, we select a suitable adjective to add.

%% file: 4_Experimental_Study.tex
\section{System}

\begin{table*}[htb]
		\centering
\begin{tabular}{|l|l|}
	\hline
	Fashion Description & Think pink but don't wear it  \\      
	Matching Quotation      & Comes with a matching slip             \\ 
	Transformed  text    & Think pink but don't match it \\        
	Transformed text + sentiment & Think gleaming pink but don't match it \\ \hline
	Fashion Description & Jewelry maybe is more expensive than clothes, but clothes are more important than jewelry  \\      
	Matching Quotation      & Accentuate the outfit with a sling bag and a bracelet             \\ 
	Transformed  text    & Jewelry maybe is more expensive than outfit, but clothes are more important than jewelry \\        
	Transformed text + sentiment & Jewelry maybe is more expensive than outfit, but stylish clothes are more important than jewelry \\ \hline
\end{tabular}
\caption{Qualitative examples illustrating the proposed approach by transforming some fashion description statements to make it more persuasive.}
\label{tab:example1}
\end{table*}

While this is a work-in-progress, we show some initial results in Table~\ref{tab:example1}, listing the input sentence, selected quotation and the transformed outputs (with and without sentiment). For input sentence, the best matching quotation is selected from the quotations corpus using the Siamese adoption of LSTM networks \cite{mueller2016siamese}. Next, suitable word replacements are identified by extracting all the nouns, verbs, adjectives, adverbs using Stanford CoreNLP parser from both the input sentence and the quote. For each selected word in the quotation, the derivationally related forms are found using the wordnet and bucketed using their POS tags (nouns, verbs, adjectives, adverbs). For example, pink\#adjective, pink\#noun, pinkify\#verb, wear\#verb, wear\#noun, etc. Next, for each of the selected word in the input sentence, the best suitable word is selected from the bucket of derivationally related forms having same POS tag as the input word. The criterion for selecting the best matching word is $matching\ score$ (discussed in the previous section). It might so happen that multiple words in the fashion sentence may match with some in the slogan, in this case, the pair which has the maximum score is selected. Using this process, we generate the transformed sentence,  shown in the third row of the table. Further, to enhance the transformation, suitable adjectives are added to the nouns. For this purpose, from a list of positive adjectives, those suitable to be used in the fashion domain were hand picked. Based on the noun in the input sentence, an appropriate adjective from the selected list is chosen and introduced into the previously transformed sentence.

%% file: 5_Working_System.tex
\section{Future Works}

In the current work, we only make use of the auxiliary information like slogans, etc to enhance the sentences related to fashion. In our observation, the fashion related sentences are tougher to deal with when compared to any news article based work predominantly discussed in the literature. Enhancing the fashion text can include idioms and similies. The existence of an ontology in this space can help  improve the transformations. The ontology can be consulted for a better word replacement in the sentences. We also plan to cover argumentative aspect of persuasion for lines that can be disputed, argued. Example "amplify the appeal with a wristwatch." could be persuaded using the argument "How could you not amplify the appeal with a wristwatch ??"

%% file: 7_Conclusion.tex
\section{Conclusion}
In this paper, we presented a system called PersuAIDE
which aims to generate persuasive text for input product description . We demonstrated the work in fashion domain where popular  expressions were used to generate creative sentences covering  memorability aspects of persuasion. Further, based on the type of sentence, sentiment features were added to slant the generated sentence. Our study can be considered
as a novel attempt to blend well known expressions
with non expressive product lines in a linguistically motivated framework that accounts for syntagmatic and paradigmatic aspects of language.

%% file: main.bbl

\begin{thebibliography}{00}


\ifx \showCODEN    \undefined \def \showCODEN     #1{\unskip}     \fi
\ifx \showDOI      \undefined \def \showDOI       #1{#1}\fi
\ifx \showISBNx    \undefined \def \showISBNx     #1{\unskip}     \fi
\ifx \showISBNxiii \undefined \def \showISBNxiii  #1{\unskip}     \fi
\ifx \showISSN     \undefined \def \showISSN      #1{\unskip}     \fi
\ifx \showLCCN     \undefined \def \showLCCN      #1{\unskip}     \fi
\ifx \shownote     \undefined \def \shownote      #1{#1}          \fi
\ifx \showarticletitle \undefined \def \showarticletitle #1{#1}   \fi
\ifx \showURL      \undefined \def \showURL       {\relax}        \fi
\providecommand\bibfield[2]{#2}
\providecommand\bibinfo[2]{#2}
\providecommand\natexlab[1]{#1}
\providecommand\showeprint[2][]{arXiv:#2}

\bibitem[\protect\citeauthoryear{De~Marneffe, Dozat, Silveira, Haverinen,
  Ginter, Nivre, and Manning}{De~Marneffe et~al\mbox{.}}{2014}]%
        {de2014universal}
\bibfield{author}{\bibinfo{person}{Marie-Catherine De~Marneffe},
  \bibinfo{person}{Timothy Dozat}, \bibinfo{person}{Natalia Silveira},
  \bibinfo{person}{Katri Haverinen}, \bibinfo{person}{Filip Ginter},
  \bibinfo{person}{Joakim Nivre}, {and} \bibinfo{person}{Christopher~D
  Manning}.} \bibinfo{year}{2014}\natexlab{}.
\newblock \showarticletitle{Universal Stanford dependencies: A cross-linguistic
  typology.}. In \bibinfo{booktitle}{{\em LREC}}, Vol.~\bibinfo{volume}{14}.
  \bibinfo{pages}{4585--92}.
\newblock


\bibitem[\protect\citeauthoryear{Gatti, {\"O}zbal, Guerini, Stock, and
  Strapparava}{Gatti et~al\mbox{.}}{2015}]%
        {gatti2015slogans}
\bibfield{author}{\bibinfo{person}{Lorenzo Gatti}, \bibinfo{person}{G{\"o}zde
  {\"O}zbal}, \bibinfo{person}{Marco Guerini}, \bibinfo{person}{Oliviero
  Stock}, {and} \bibinfo{person}{Carlo Strapparava}.}
  \bibinfo{year}{2015}\natexlab{}.
\newblock \showarticletitle{Slogans Are Not Forever: Adapting Linguistic
  Expressions to the News.}. In \bibinfo{booktitle}{{\em IJCAI}}.
  \bibinfo{pages}{2452--2458}.
\newblock


\bibitem[\protect\citeauthoryear{Gatti, Ozbal, Guerini, Stock, and
  Strapparava}{Gatti et~al\mbox{.}}{2016}]%
        {gatti2016heady}
\bibfield{author}{\bibinfo{person}{Lorenzo Gatti}, \bibinfo{person}{Gozde
  Ozbal}, \bibinfo{person}{Marco Guerini}, \bibinfo{person}{Oliviero Stock},
  {and} \bibinfo{person}{Carlo Strapparava}.} \bibinfo{year}{2016}\natexlab{}.
\newblock \showarticletitle{Heady-Lines: A Creative Generator Of Newspaper
  Headlines}. In \bibinfo{booktitle}{{\em Companion Publication of the 21st
  International Conference on Intelligent User Interfaces}}. ACM,
  \bibinfo{pages}{79--83}.
\newblock


\bibitem[\protect\citeauthoryear{Guerini, {\"O}zbal, and Strapparava}{Guerini
  et~al\mbox{.}}{2015}]%
        {guerini2015echoes}
\bibfield{author}{\bibinfo{person}{Marco Guerini}, \bibinfo{person}{G{\"o}zde
  {\"O}zbal}, {and} \bibinfo{person}{Carlo Strapparava}.}
  \bibinfo{year}{2015}\natexlab{}.
\newblock \showarticletitle{Echoes of persuasion: The effect of euphony in
  persuasive communication}.
\newblock \bibinfo{journal}{{\em arXiv preprint arXiv:1508.05817\/}}
  (\bibinfo{year}{2015}).
\newblock


\bibitem[\protect\citeauthoryear{Guerini, Strapparava, and Stock}{Guerini
  et~al\mbox{.}}{2011}]%
        {guerini2011slanting}
\bibfield{author}{\bibinfo{person}{Marco Guerini}, \bibinfo{person}{Carlo
  Strapparava}, {and} \bibinfo{person}{Oliviero Stock}.}
  \bibinfo{year}{2011}\natexlab{}.
\newblock \showarticletitle{Slanting existing text with Valentino}. In
  \bibinfo{booktitle}{{\em Proceedings of the 16th international conference on
  Intelligent user interfaces}}. ACM, \bibinfo{pages}{439--440}.
\newblock


\bibitem[\protect\citeauthoryear{Lynch}{Lynch}{}]%
        {lynchevery}
\bibfield{author}{\bibinfo{person}{Gerard Lynch}.}
\newblock \showarticletitle{Every Word You Set: Simulating the cognitive
  process of linguistic creativity with the PUNdit system}.
\newblock  (\bibinfo{year}{????}).
\newblock


\bibitem[\protect\citeauthoryear{Mueller and Thyagarajan}{Mueller and
  Thyagarajan}{2016}]%
        {mueller2016siamese}
\bibfield{author}{\bibinfo{person}{Jonas Mueller} {and} \bibinfo{person}{Aditya
  Thyagarajan}.} \bibinfo{year}{2016}\natexlab{}.
\newblock \showarticletitle{Siamese Recurrent Architectures for Learning
  Sentence Similarity.}. In \bibinfo{booktitle}{{\em AAAI}}.
  \bibinfo{pages}{2786--2792}.
\newblock


\bibitem[\protect\citeauthoryear{{\"O}zbal, Pighin, and Strapparava}{{\"O}zbal
  et~al\mbox{.}}{2013}]%
        {ozbal2013brainsup}
\bibfield{author}{\bibinfo{person}{G{\"o}zde {\"O}zbal},
  \bibinfo{person}{Daniele Pighin}, {and} \bibinfo{person}{Carlo Strapparava}.}
  \bibinfo{year}{2013}\natexlab{}.
\newblock \showarticletitle{BRAINSUP: Brainstorming Support for Creative
  Sentence Generation.}. In \bibinfo{booktitle}{{\em ACL (1)}}.
  \bibinfo{pages}{1446--1455}.
\newblock


\bibitem[\protect\citeauthoryear{Pang, Lee, et~al\mbox{.}}{Pang
  et~al\mbox{.}}{2008}]%
        {pang2008opinion}
\bibfield{author}{\bibinfo{person}{Bo Pang}, \bibinfo{person}{Lillian Lee},
  {et~al\mbox{.}}} \bibinfo{year}{2008}\natexlab{}.
\newblock \showarticletitle{Opinion mining and sentiment analysis}.
\newblock \bibinfo{journal}{{\em Foundations and Trends{\textregistered} in
  Information Retrieval\/}} \bibinfo{volume}{2}, \bibinfo{number}{1--2}
  (\bibinfo{year}{2008}), \bibinfo{pages}{1--135}.
\newblock


\bibitem[\protect\citeauthoryear{Serban, Garc{\'\i}a-Dur{\'a}n, Gulcehre, Ahn,
  Chandar, Courville, and Bengio}{Serban et~al\mbox{.}}{2016}]%
        {serban2016generating}
\bibfield{author}{\bibinfo{person}{Iulian~Vlad Serban},
  \bibinfo{person}{Alberto Garc{\'\i}a-Dur{\'a}n}, \bibinfo{person}{Caglar
  Gulcehre}, \bibinfo{person}{Sungjin Ahn}, \bibinfo{person}{Sarath Chandar},
  \bibinfo{person}{Aaron Courville}, {and} \bibinfo{person}{Yoshua Bengio}.}
  \bibinfo{year}{2016}\natexlab{}.
\newblock \showarticletitle{Generating factoid questions with recurrent neural
  networks: The 30m factoid question-answer corpus}.
\newblock \bibinfo{journal}{{\em arXiv preprint arXiv:1603.06807\/}}
  (\bibinfo{year}{2016}).
\newblock


\bibitem[\protect\citeauthoryear{Toivanen, Toivonen, Valitutti, Gross,
  et~al\mbox{.}}{Toivanen et~al\mbox{.}}{2012}]%
        {toivanen2012corpus}
\bibfield{author}{\bibinfo{person}{Jukka Toivanen}, \bibinfo{person}{Hannu
  Toivonen}, \bibinfo{person}{Alessandro Valitutti}, \bibinfo{person}{Oskar
  Gross}, {et~al\mbox{.}}} \bibinfo{year}{2012}\natexlab{}.
\newblock \showarticletitle{Corpus-based generation of content and form in
  poetry}. In \bibinfo{booktitle}{{\em Proceedings of the Third International
  Conference on Computational Creativity}}.
\newblock


\bibitem[\protect\citeauthoryear{Zheng, Si, Chang, and Zhu}{Zheng
  et~al\mbox{.}}{2011}]%
        {37566}
\bibfield{author}{\bibinfo{person}{Zhicheng Zheng}, \bibinfo{person}{Xiance
  Si}, \bibinfo{person}{Edward~Y. Chang}, {and} \bibinfo{person}{Xiaoyan Zhu}.}
  \bibinfo{year}{2011}\natexlab{}.
\newblock \showarticletitle{K2Q: Generating Natural Language Questions from
  Keywords with User Refinements}. In \bibinfo{booktitle}{{\em Proceedings of
  the 5th International Joint Conference on Natural Language Processing}}.
  \bibinfo{pages}{947–955}.
\newblock
\showURL{%
\url{http://aclweb.org/anthology-new/I/I11/I11-1106.pdf}}


\end{thebibliography}
